\documentclass[final,5p,times,twocolumn]{elsarticle}
\usepackage{graphicx}
\usepackage{amsmath}
\usepackage{amssymb}
\usepackage{booktabs}
\usepackage{mathtools}
\usepackage{fancyhdr}
\DeclareMathOperator*{\argminA}{arg\,min}
\usepackage{flexisym}
\usepackage{latexsym}
\usepackage{multirow}
\usepackage{bm}
\usepackage{breqn}
\usepackage{color}
\newcommand{\etal}{\textit{et al}. }

\usepackage[pagebackref=true,breaklinks=true,colorlinks,bookmarks=false]{hyperref}
\journal{Neurocomputing}
\begin{document}

\begin{frontmatter}

\pagestyle{fancy}
\title{Deep Model Compression Based on the Training History
}

% \begin{frontmatter}
% \pagestyle{fancy}
% \title{Deep Model Compression Based on the Training History}
\author{S.H.Shabbeer Basha$^1$,Mohammad Farazuddin$^1$,Viswanath Pulabaigari$^1$,
Shiv Ram Dubey$^2$,Snehasis Mukherjee$^3$}

\address{{$^1$Indian Institute of Information Technology Sri City, Chittor, India-517646.\\
$^2$Indian Institute of Information Technology, Allahabad, Uttar Pradesh- 211015, India. \\
$^3$Shiv Nadar University, Uttar Pradesh, India. \\
Email: \{shabbeer.sh, farazuddin.m17, viswanath.p\}@iiits.in}, srdubey@iiita.ac.in, snehasis.mukherjee@snu.edu.in}

% \author{S.H.Shabbeer Basha$^1$, Mohammad Farazuddin$^1$, Viswanath P$^1$, Shiv Ram Dubey$^2$, Snehasis Mukherjee$^3$}
% \address{
% $^1$Computer Vision Group, Indian Institute of Information Technology, Sri City, Chittoor, Andhra Pradesh- 517646, India.\\
% $^2$Computer Vision and Biometrics Laboratory, Indian Institute of Information Technology, Allahabad, Uttar Pradesh- 211015, India. \\
% $^3$Department of Computer Science and Engineering, Shiv Nadar University,  Uttar Pradesh, India. 
% \\[\bigskipamount]
% {\texttt{\{shabbeer.sh, farazuddin.m17, viswanath.p\}@iiits.in, srdubey@iiita.ac.in, snehasis.mukherjee@snu.edu.in}}}
% \author{Mohammad Farazuddin, Viswanath P, Shiv Ram Dubey, \\
% Indian Institute of Information Technology, Sri City, \\
% Andhra Pradesh-517646, India \\
% {\tt\small shabbeer.sh,farazuddin.m17,viswanath.p,srdubey@iiits.in}
% \and
% Snehasis Mukherjee\\
% Shiv Nadar University\\
% Greater Noida, Uttar Pradesh-201314, India\\
% {\tt\small snehasis.mukherjee@snu.edu.in}
% }
% \maketitle
%%%%%%%%% ABSTRACT
\begin{abstract}
   Deep Convolutional Neural Networks (DCNNs) have shown promising performances in several visual recognition problems which motivated the researchers to propose popular architectures such as LeNet, AlexNet, VGGNet, ResNet, and many more. These architectures come at a cost of high computational complexity and parameter storage. To get rid of storage and computational complexity, deep model compression methods have been evolved. We propose a ``History Based Filter Pruning (HBFP)" method that utilizes network training history for filter pruning. Specifically, we prune the redundant filters by observing similar patterns in the filter's $\ell_{1}$-norms (absolute sum of weights) over the training epochs. We iteratively prune the redundant filters of a CNN in three steps. First, we train the model and select the filter pairs with redundant filters in each pair. Next, we optimize the network to ensure an increased measure of similarity between the filters in a pair. 
%   It decreases the information loss if one of the filter from a pair is pruned.
  This optimization of the network facilitates us to prune one filter from each pair based on its importance without much information loss. Finally, we retrain the network to regain the performance, which is dropped due to filter pruning. We test our approach on popular architectures such as LeNet-5 on MNIST dataset; VGG-16, ResNet-56, and ResNet-110 on CIFAR-10 dataset, and ResNet-50 on ImageNet. The proposed pruning method outperforms the state-of-the-art in terms of FLOPs reduction (floating-point operations) by $\textbf{97.98 \%}$, $\textbf{83.42 \%}$, $\textbf{78.43 \%}$, $\textbf{74.95 \%}$, and $\textbf{75.45 \%}$  for LeNet-5, VGG-16, ResNet-56, ResNet-110, and ResNet-50, respectively, while maintaining the less error rate.
\end{abstract}
\end{frontmatter}
% \end{frontmatter}

\section{Introduction}
\label{sec:intro}
In recent years, Convolutional Neural Networks (CNN) have gained significant attention from researchers, especially for visual recognition tasks, due to their impeccable performance in several tasks including object recognition and detection \cite{lecun2015deep}, speech recognition \cite{fayek2017evaluating}. The wide usage of deep CNNs in numerous applications (especially in vision) creates an increasing demand for memory (parameter storage) and computation. To address this key issue, various attempts have been made in the literature. One such attempt focuses on training the deep CNNs with limited data \cite{finn2018probabilistic,kumar2018generalized,yoon2018bayesian}. Another line of research has shown better performance by reducing the overhead of computational power and memory storage, which mainly focuses on model compression by two approaches: pruning connections \cite{han2015deep,wu2018blockdrop} and pruning filters \cite{denton2014exploiting,he2017channel,li2016pruning}.

\begin{figure*}[!t]
    \centering
    \includegraphics[width=\textwidth]{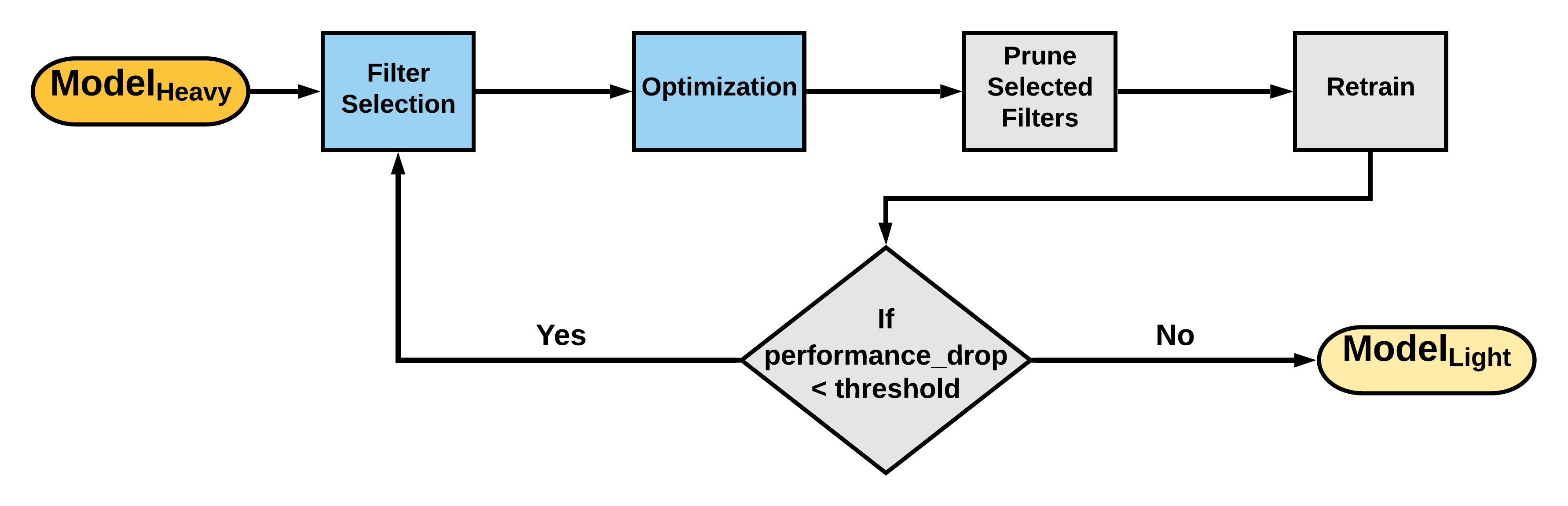}
    \caption{Initially, we start with a heavier model, after identifying the redundant filters, instead of pruning the filters naively at this stage, we optimize the model to minimize the complimentary information loss that will occur due to filter pruning. After optimizing the network with the custom regularizer, one filter from each redundant filter pair is pruned. Later the network is re-trained to regain the classification performance. This process is repeated iteratively to obtain a lightweight CNN.}
    \label{fig:proposed_method_broadview}
\end{figure*}

Typically, increasing the size (requires more storage space) of a deep neural network makes deploying the model difficult on low-end (resource constraint) devices such as mobile devices and embedded systems. For example, VGG-16 \cite{simonyan2014very} has 138.34 million parameters which require storage space of more than 500 MB. To reduce the resource overhead of deep CNNs, many attempts have been made to prune less-important connections and filters from a CNN which results in architectures with compressed design. Most of the works on deep model compression can be broadly categorized into four classes \cite{liang2021pruning}. The first class \cite{chen2015compressing,han2015deep} of methods are focused on introducing sparsity into the model parameters. The second class \cite{louizos2017bayesian,rastegari2016xnor,polino2018model} of methods are aimed at quantization-based pruning. The third class of methods are dedicated to compressing the networks using filter decomposition \cite{li2019learning,ding2019centripetal,li2020group}. The fourth class of methods are focused on pruning unimportant filters \cite{abbasi2017structural,denton2014exploiting,zhang2015efficient}. The proposed training History Based Filter Pruning (HBFP) method belongs to the fourth category.  

In general, filter pruning methods require some metric to calculate the importance of a filter. Several metrics have been proposed to calculate the importance of a filter. For instance, Abbasi \etal \cite{abbasi2017structural} employed a brute-force technique to prune the filters sequentially that contribute less to the classification performance. However, the brute-force technique is inefficient while dealing with large neural networks, such as AlexNet \cite{krizhevsky2012imagenet} and VGG-16 \cite{simonyan2014very}. Li \etal \cite{li2016pruning} pruned the unimportant filters based on their $\ell_1$-norm. They assumed that the filters with a high $\ell_1$-norm are most probably important and will have a larger influence on the relevance of the generated feature map.

In this paper, we introduce a novel method for pruning the redundant filters based on the training history. We iteratively prune redundant filters from a CNN in three stages. First, we select some ($M \%$ of) filter pairs as redundant for which the sum of the absolute value of the difference between the filter's $\ell_1$-norm over training epochs is minimum. Next, instead of pruning the filters directly, we reduce the difference between the filter's $\ell_1$-norm in respective epochs (which we call optimization) to minimize the complimentary information loss and then prune one filter from each pair based on its magnitude. Finally, we fine-tune (re-train) the network to gain the classification performance, which is decreased due to filter pruning. We consider $\ell_1$-norm of filters across the training epochs to select and prune the redundant filters due to its simplicity. However, our method can be adapted to any other metric like $\ell_2$-norm, Cosine Similarity, and so on. The high-level view of the proposed method is outlined in Fig.\ref{fig:proposed_method_broadview}.  

The remaining paper is organized as follows: the related works are compiled in Section \ref{sec:related_work}; Preliminaries and notations are presented in Section \ref{sec:preliminaries};
The proposed History Based Filter Pruning (HBFP) method is explained in Section \ref{sec:proposed_method}; The experimental pruning results are examined in Section \ref{sec:experimental_results} along with the analysis and discussion; Finally, the concluding remarks are made in Section \ref{sec:conclusion} along with the future directives.
\section{Related Works}
\label{sec:related_work}
We illustrate the efforts found in the published literature for deep model compression, separately by classifying the approaches into four major categories mentioned in the Introduction section.

\subsection{Connection Pruning}
Connection pruning methods induce sparsity into the neural network. A simple approach is to prune the connections with unimportant weights (parameters) \cite{xie2019learning}. However, this method requires quantifying the significance of the parameters. In this direction, Lecun \etal \cite{lecun1990optimal} and Hassibil \etal \cite{hassibi1993second} have utilized second-order derivative information to quantify the importance of network connections (parameters). However, computing second-order derivatives of all the connections is expensive. Chen \etal \cite{chen2015compressing} used a low-cost hash function to group the weights into a single bucket such that weights in the same bucket have roughly the same parameter value. Hu \etal \cite{hu2016network} introduced a network trimming approach that iteratively prunes the zero-activation neurons. Wu \etal \cite{wu2018blockdrop} proposed a method called BlockDrop to dynamically learn which layers to execute during the inference to reduce the total computation time. Han \etal \cite{han2015deep} developed a pruning technique based on the absolute value of the parameter. In \cite{han2015deep}, the parameters with the absolute value below a certain threshold are fixed to zero. These pruning methods are suggested in the scenarios where the majority of the network parameters belong to Fully Connected (FC) layers. For deep models such as ResNet \cite{he2016deep} and DenseNet \cite{huang2017densely}, these types of pruning methods might not be suitable. However, the use of modern deep learning models for different applications is a recent trend.

\subsection{Weight Quantization}
The weight (Parameter) quantization method is found in the literature used extensively for deep model compression \cite{}. Han \etal \cite{han2015deep} compressed the deep CNNs by integrating connection pruning, quantization, and Huffman coding. Similarly, Tung \etal \cite{tung2018clip} combined pruning and weight quantization for deep model compression. Floating-point quantization is performed in \cite{miao2017towards} for creating efficient deep neural networks. Binarization \cite{rastegari2016xnor} is another popular quantization technique used for model compression in which each floating-point value is mapped to a binary value. Bayesian approximation methods \cite{louizos2017bayesian} are used for deep model quantization. The weight quantization-based methods aim to speed up the execution process by reducing the complexity of number representation and arithmetic \& logical operations. However, these methods require the support of special hardware to capture the benefit of network compression. Recently, two-way model compression schemes are proposed in \cite{ruan2020edp,ma2021non} in which both filter decomposition and pruning are performed.

\subsection{Filter Decomposition}
As reported in \cite{denil2013predicting}, deep neural networks are over-parameterized that indicates that the parameters of a layer can be recovered from a subset of the actual parameters that belong to the same layer. Motivated by this work, many low-rank filter decomposition works have been evolved \cite{li2019learning,ding2019centripetal,li2020group}. Unlike filter pruning, which aims at pruning the unimportant filters, these methods decrease the computational cost of the network. In this direction, Denton \etal \cite{denton2014exploiting} utilized the linear structure of CNNs to find a suitable low-rank approximation for the parameters by allowing minimal loss to the network performance. Zhang \etal \cite{zhang2015efficient} made use of subsequent non-linear units for learning low-rank filter decomposition to speed up the learning process. Lin \etal \cite{lin2018holistic} introduced a low-rank decomposition method to decrease the redundant features corresponding to convolutional kernels and dense layer matrices.

\subsection{Filter Pruning}
Compared to other network pruning methods, filter pruning methods are generic, which do not require the support of any special software/hardware. Due to this reason, filter pruning methods have gained popularity among researchers in recent years. In general, filter pruning methods \cite{lin2020hrank,he2018soft,he2017channel,luo2017thinet,zhang2021filter} compute the importance of filters so that unimportant filters can be pruned from the model. In filter-pruning methods, after each iteration, retraining is required to regain the classification performance, which is dropped due to pruning the filters. Abbasi \etal \cite{abbasi2017structural} proposed a greedy-based compression scheme for filter pruning. Similarly, Li \etal \cite{li2016pruning} employed a greedy approach to prune the filters with less filter norm. In \cite{zhao2019variational}, redundant channels are investigated based on the distribution of channel parameters. Ayinde \etal \cite{ayinde2019redundant} utilized relative cosine distance among the filters for filter pruning. Ding \etal \cite{ding2018auto} proposed an auto-balanced method to transfer the representation capacity of a convolutional layer to a fraction of filters belong to the same layer. Other methods such as Taylor expansion \cite{molchanov2016pruning}, low-rank approximation \cite{denton2014exploiting,jaderberg2014speeding,zhang2015efficient}, group-wise sparsity \cite{lebedev2016fast,wen2016learning,zhou2016less,alvarez2016learning}, and many more are employed to prune the filters from deep neural networks. Zhang \etal \cite{zhang2021structadmm} introduced a unified framework that can be used to induce sparsity at various granularities like filter-wise, channel-wise, and shape-wise sparsity. Recently, Lin \etal \cite{lin2020hrank} proposed a filter pruning method based on the rank of feature maps in each layer such that the filters contributing to low-rank feature maps can be pruned. Chen \etal \cite{chen2020dynamical} developed a filter pruning method that dynamically prunes channels/filters during network training rather than iteratively pruning and re-training. Song \etal \cite{song2020sp} introduced SP-GAN, a self-growing and pruning Generative Adversarial Network for image generation, in which, Euclidean distance between each pair of filters is used as the metric for filter pruning. Later, a filter is randomly selected for pruning from the filter pairs having the least Euclidean distance. He \etal \cite{he2019asymptotic} proposed a soft-filter pruning method that allows filters to be updated during model training which results in better performance. Recently, Wang \etal \cite{wang2021filter} proposed a filter pruning method computes the importance of filters based on entropy of feature maps. Later, low-ranked filters are pruned from the model.

% Very recently, Lin \etal \cite{lin2020hrank} proposed a pruning technique based on the filter-correlations which prunes the redundant convolution filters. They pruned one filter from each filter-pair after increasing the similarity between the filters within a pair to reduce the information loss incurred due to pruning.     

\begin{figure*}[!t]
    \centering
    \includegraphics[width=\textwidth]{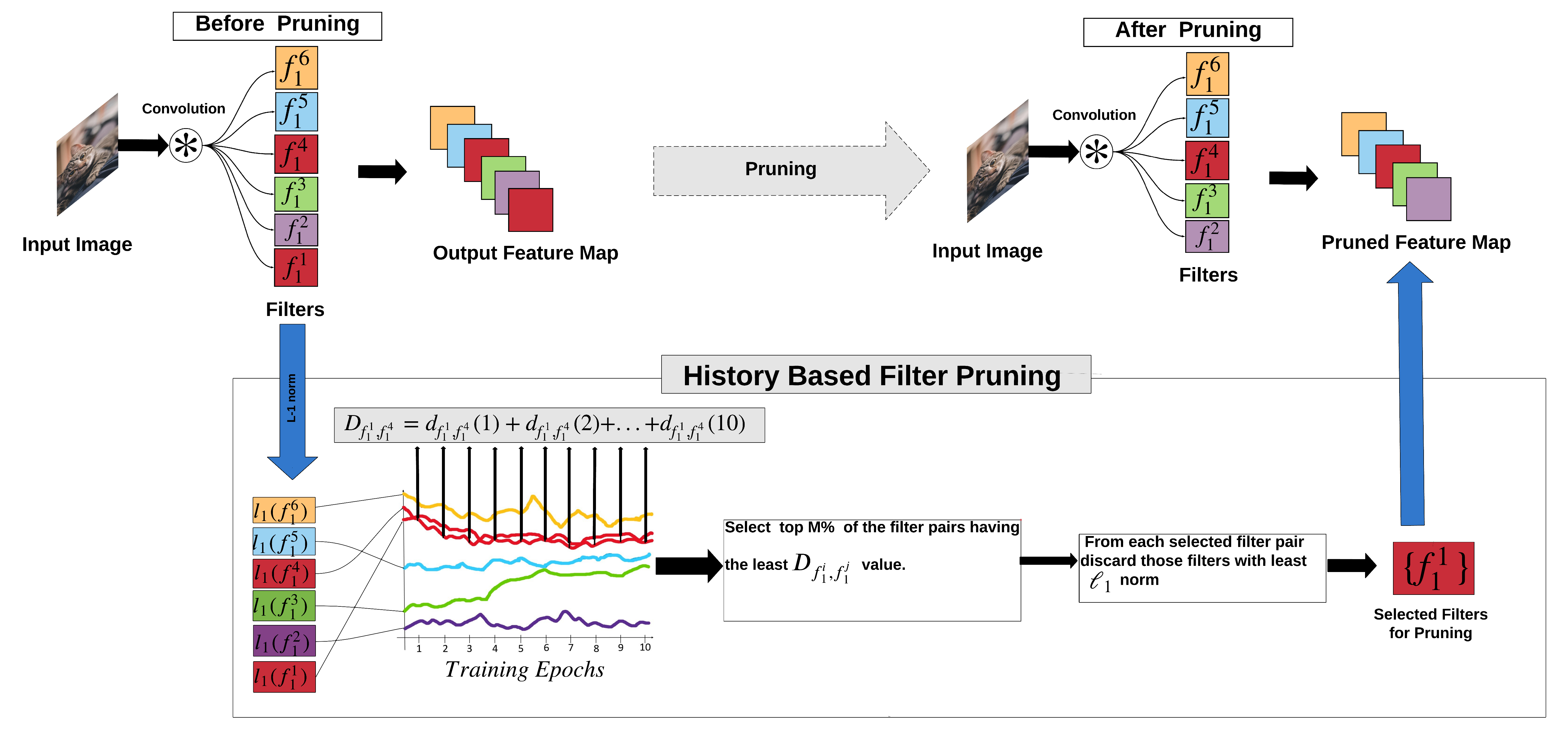}
    \caption{The overview of the proposed pruning method. At each iteration, we compute the sum of differences $D_{f_{k}^{i}, f_{k}^{j}}$ of all filter pairs. In filter selection stage, we choose $M \%$ of filter pairs for which $D_{f_{k}^{i}, f_{k}^{j}}$ is  minimum. Here, filters $f_1^1$ and $f_1^4$ have the least $D_{f_{k}^{i}, f_{k}^{j}}$ (absolute point-to-point difference), so we select them as a pair and discard one from the pair with least magnitude. The feature maps corresponding to the pruned filters get dropped.}
    \label{fig:proposed_method}
\end{figure*}

Apart from the neural network compression methods discussed above, there are methods in which a small model (student) mimics the behavior of a large model (teacher). This paradigm is popularly known as Knowledge Distillation in the literature \cite{gou2021knowledge}. Recently, EDropout is proposed in \cite{salehinejad2021edropout} which uses a binary state pruning vector to prune the filters/units from convolutional, dense layers. A comprehensive survey on deep model compression is presented by Cheng \etal \cite{chen2015compressing}.   

Most of the filter pruning methods discussed so far, prune the unimportant filters. However, these methods may not remove the filters that are consistently redundant throughout the network training. We propose a novel filter pruning technique that utilizes the training history to find the filters to be pruned. Moreover, in contrast to other classes of pruning methods, the proposed filter pruning method does not require the support of any additional software/hardware. In brief, the contributions of this research can be summarized as follows,
\begin{itemize}
    \item We propose a novel method for pruning filters from convolutional layers based on the training history of a deep neural network. The proposed method facilitates the identification of stable and redundant filters throughout the training that can have a negligible effect on performance after pruning.
    \item We introduce an optimization step (custom regularizer) to reduce the information loss incurred due to filter pruning. It is achieved by increasing the redundancy level of selected filters for pruning.
    \item To establish the significance of the proposed pruning method, experiments are conducted on benchmark CNNs like LeNet-5 \cite{lecun1998gradient}, VGG-16 \cite{simonyan2014very}, ResNet-50, ResNet-56, and ResNet-110 \cite{he2016deep}. The validation of the proposed pruning method is performed over three benchmark classification datasets, including MNIST, CIFAR-10, and ImageNet.
\end{itemize}

% Also, note that the other pruning methods such as low-rank approximation methods can be integrated into our method (to decompose dense layers) to obtain a better-compressed model. \textcolor{red}{(Are we showing any such results... if not better not to mention....)}

\section{Preliminaries and Notations}
\label{sec:preliminaries}
In this section, we discuss the background details like computing the Floating Point Operations (FLOPs) involved in a CNN and the notations used in this paper.

\subsection{Calculating Floating Point Operations}
To compare the performance of various CNN models, we primarily use the accuracy on the validation set as a metric to compare which model is the most accurate. However, when there are constraints on computational resources, we use the number of Floating Point Operations (FLOPs) as a metric to compare which model is more efficient. We use the terms ``Heavy" and ``Light" to represent the models with a higher and lower number of FLOPs, respectively. For a given input feature map, the number of FLOPs involved in a convolutional layer $L_{i}$, i.e., ($FLOP_{conv}(L_{i})$), is computed as follows, 
\begin{equation}
    FLOP_{conv}(L_{i}) = F * F * C_{in} * H_{out} * W_{out} * C_{out}.
  \label{eq:flops_conv}
\end{equation}
Here, $F*F$ is the spatial dimension of the filter, $C_{in}$ is the number of input channels of the input feature map, $H_{out}$ and $W_{out}$ are the height and width of the output feature map, and $C_{out}$ is the number of channels in the output feature map. Similarly, for a given input feature map, the number of FLOPs for a Fully Connected (FC) or dense layer $D_{i}$, i.e., ($FLOP_{fc}(D_{i}))$, is given as,
\begin{equation}
    FLOP_{fc}(D_{i}) = C_{in} * C_{out}.
  \label{eq:flops_fc}
\end{equation}
For a model with $K$ convolutional layers and $N$ fully-connected layers, the total number of FLOPs is calculated as,
\begin{equation}
    FLOP_{total} = \sum_{i=1}^{K}FLOP_{conv}(L_{i}) + \sum_{j=1}^{N}FLOP_{fc}(D_{j}).
  \label{eq:flops_total}
\end{equation}

\subsection{Notations}
Consider a convolutional layer $L_i$ of a CNN, which has $n$ filters, i.e., \{$f^1, f^2, f^3, ..., f^n$\}. Any two filters $f^i$, $f^j$ belong to the $k^{th}$ layer of a CNN are denoted as $f_{k}^{i}$, $f_{k}^{j}$, respectively. For instance, if the filter $f_{k}^{i}$ is of dimension $3\times3\times3$ then it consists of 27 parameters, i.e., $\{w_{k,1}^{i}, w_{k,2}^{i}, w_{k,3}^{i}, ..., w_{k,27}^{i}\}$. Here `$k$' represents the index of convolutional layer and `$i$' denotes the filter index.

Initially, our method computes the $\ell_1$-norm of each filter within the same convolutional layer using the formula given in Eq. \ref{eq:l_{1}-norm}. For example the $\ell_1$-norm of filter $f_{k}^{i}$ is computed as follows, 
\begin{equation}
    \ell_1(f_{k}^{i}) = \|f^{i}_{k}\|_{1} = 
    \sum_{p=1}^{27}|{w_{k,p}^i}|,
  \label{eq:l_{1}-norm}
\end{equation}
where $p$ is the number of parameters in the filter $f_k^i$. 

% Here, we assume the filter $f_{k}^{i}$ is of dimension $3\times3$.  

Next, we compute the absolute difference between the $\ell_1$-norms of each filter pair at every epoch as shown in Eq. \ref{eq:difference}. The absolute difference between $\ell_1(f_{k}^{i})$ and $\ell_1(f_{k}^{j})$ is computed as follows,

\begin{equation}
    d_{f_{k}^{i}, f_{k}^{j}}(t) = {|{\ell_1(f_{k}^{i})-\ell_1(f_{k}^{j})}|}.
  \label{eq:difference}
\end{equation}
Here, `$t$' indicates the epoch number. Then, this difference is summed over all the epochs and denoted by $D_{f_{k}^{i}, f_{k}^{j}}$. The sum of differences of filter pairs over the training epochs is considered as the metric for filter pruning and given as,
\begin{equation}
    D_{f_{k}^{i}, f_{k}^{j}} =  \sum_{t=1}^{N}d_{f_{k}^{i},f_{k}^{j}}(t),
  \label{eq:finaldifference}
\end{equation}
where `$N$' indicates the maximum number of epochs used for training the networks. 

Our method computes the sum of differences of pairs $D_{f_{k}^{i}, f_{k}^{j}}$ value for $n_{C_{2}}$ filter pairs (assuming a convolutional layer has $n$ filters). The total difference $D_{f_{k}^{i}, f_{k}^{j}}$ which is summed over all the epochs is used for filter selection. More concretely, the top $M\%$ of the filter pairs (from $n_{C_{2}}$ pairs in the same layer) with the least $D_{f_{k}^{i}, f_{k}^{j}}$ value are formed as redundant filter pairs which represent roughly the same information. The overview of the proposed HBFP method is presented in Fig. \ref{fig:proposed_method}. The proposed  History Based Filter Pruning (HBFP) method involves two key steps, i) Filter selection and ii) Optimization which are presented in the next section.

\begin{table}[!t]
\centering
\caption{The pruning results of LeNet-5 on MNIST dataset. The rows corresponding to HBFP-I, HBFP-II, HBFP-III, and HBFP-IV indicate the pruning results of the last four iterations of the proposed method. Here * indicates the reproduced results. The results are arranged in the increasing order of pruned FLOPs reduction \%.    
% We conduct experiments for three times and report the mean $\pm$ standard deviation. The reported results are taken from the respective research article.
% The best performing metrics are highlighted in bold. 
}
\resizebox{\columnwidth}{!}{%
\begin{tabular}{|l|l|l|l|}
\hline
Method                  & r1, r2  &  \begin{tabular}[c]{@{}l@{}}Top-1$\%$ \\ {Error} \end{tabular}           &  \begin{tabular}[c]{@{}l@{}}Remaining \\ FLOPs \\ {(Drop \%)} \end{tabular}   \\ \hline

Baseline  & 20,50 & 0.83 & 4.4M  (0.0\%) \\  

Sparse-VD  \cite{molchanov2017variational}                          & -      & 0.75   &   2.0M (54.34\%)                                                                                                    \\ 

SBP \cite{neklyudov2017structured} & -    & 0.86                                      &   0.41M (90.47\%)                    \\

SSL-3 \cite{wen2016learning} & 3,12 & 1.00 & 0.28M (93.42\%) \\

\textbf{HBFP-I (Ours)}               & 4,5  &  0.98  & 0.19M (95.57\%) \\ 

GAL \cite{lin2019towards} & 2,15
& 1.01 & 0.10M (95.60\%) \\

\textbf{HBFP-II (Ours)}                 & 3,5  & 1.08   & 0.15M (96.41\%) \\ 

Auto balanced \cite{ding2018auto}                                & 3,5          & 2.21            & 0.15M  (96.41\%)          \\ 
% $\ell_1$-norm \cite{li2016pruning}                         & $2.06$          & 6.60            & 34.20          \\ \hline

% Hinge \cite{li2020group} & 93.59 & 122.5M (60.93\%) & 11.99M (19.95\%) \\

% HRank \cite{lin2020hrank} & 91.23 & 73.7M (76.5\%) & 1.78M (92.0\%) \\

% FO \cite{qin2018functionality}          & $1.85$          & 6.70            & 44.10          \\ \hline
% PFF \cite{}

\textbf{HBFP-III (Ours)}             & 3,4  & 1.20    &  0.13M (96.84\%) \\ 

CFP \cite{singh2020leveraging}          & 2,3 &  $1.77$          & 0.08M (97.98\%)                      \\ 
CFP \cite{singh2020leveraging}$^*$          & 2,3 &  $2.61$          & 0.08M (97.98\%)                      \\ 

% Auto-balanced FP \cite{ding2018auto}          & $0.58$          & 7.13            & 81.39          \\ \hline

\textbf{HBFP-IV (Ours)}                 & 2,3  & 1.40   & 0.08M (97.98\%) \\ \hline

\end{tabular}
}
\label{tab:lenet_results}
\end{table}

\section{Proposed Training History Based Filter Pruning (HBFP) Method}
\label{sec:proposed_method}
Our pruning method aims at making a deep neural network computationally efficient. This is achieved by pruning the redundant filters whose removal do not cause much hindrance to the classification performance. We identify the redundant filters by observing similar patterns in the weights (parameters) of filters during the network training, which we refer to as the \mbox{network's} training history. We start with a pre-trained CNN model. During the network training, we observe and form pairs of filters whose weights follow a similar trend over the training epochs. In each iteration, we pick some ($M \%$) of the top filter pairs with high similarity (based on the $D$ value computed using Eq. \ref{eq:finaldifference}, low $D$ value means high similarity). Instead of pruning the filters at this stage, we increase the similarity between the filters that belong to the selected filter pairs by introducing an optimization step. This optimization is achieved with a custom regularizer whose objective is to minimize the difference between the filter's norms (belong to a filter pair) at each epoch. After optimization, one filter from each pair is discarded (pruned). From a filter pair, we prune one filter based on the criteria employed in \cite{li2016pruning}, i.e., the filters with a higher $\ell_1$-norm are more important. Finally, to recover the model from the performance drop which is incurred due to filter pruning, we retrain the pruned model. This process corresponds to one iteration of the proposed pruning method which is demonstrated in Fig.\ref{fig:proposed_method_broadview}. This whole process is repeated until the model's performance drops below a certain threshold. Our main contributions are made specifically in the filter selection and optimization steps.

\subsection{Filter Selection}
In the beginning, our method takes a heavy-weight CNN and selects the top $M\%$ of filter pairs from each convolutional layer for which the difference $D$ computed in Eq. \ref{eq:finaldifference} is minimum. More concretely, the filter pair having the least $D_{f_{k}^{i}, f_{k}^{j}}$ value is formed as the first redundant filter pair. Similarly, the next pair having the second least $D_{f_{k}^{i}, f_{k}^{j}}$ value is formed as another filter pair and so on. Likewise, in each iteration, $M \%$ of filter pairs from each convolutional layer are selected as redundant which are further considered for optimization. Let us define two more terms that are used in the proposed pruning method. ``Qualified-for-pruning ($Q_i$)" and ``Already Pruned ($P_i$)". Here $Q_i$ represents the set of filter pairs that are ready (selected) for pruning from a convolutional layer $L_i$. Whereas, $P_i$ indicates the filters that are pruned from the network (one filter from each pair of $Q_i$). Hence, if $M\%$ of filter pairs are chosen in $Q_i$, then $|P|=M \%$, i.e., from each convolutional layer $M \%$ of filters are pruned in every iteration by the proposed method.

\subsection{Optimization}
Singh \etal \cite{singh2020leveraging} reported that introducing an optimization with a custom regularizer decreases the information loss incurred due to filter pruning. Motivated by this work, we add a new regularizer to the objective function to reduce the difference $d_{f_{k}^{i}, f_{k}^{j}}(t)$ between the filters belonging to $Q_i$ at each epoch during network training, i.e., increasing the similarity between the filters belong to the same pair. Let $C(W)$ be the objective function (Cross-entropy loss function) of the deep convolutional neural network with $W$ as the network parameters. To minimize the information loss and to maximize the regularization capability of the network, we employ a custom regularizer to the objective function, which is given as follows:
\begin{equation}
C_1 = \exp{\Bigg (\sum_{f_{k}^{i}, f_{k}^{j}\in Q_{i}}d_{f_{k}^{i}, f_{k}^{j}}(t)\Bigg )},
    \label{eq:regularizer}
\end{equation}
where $t$ denotes the epoch number and $t \in 1,2,3,..., N$ (assuming we train the model for $N$ epochs). With this new regularizer, the final objective of the proposed HBFP method is given by,

\begin{equation}
W = \argminA_W \Bigg (C(W)+\lambda*C_1 \Bigg ),
\label{eq:objective}    
\end{equation}
where $\lambda$ is the regularizer term which is a hyperparameter. Optimizing the Eq. \ref{eq:objective} decreases the difference $d_{f_{k}^{i}, f_{k}^{j}}$ between the filter pairs that belong to the set $Q_i$ at every epoch without affecting the model's performance much.

\subsection{Pruning and Re-training}
% This process facilitates us to prune one filter from each pair based on its magnitude as in \cite{li2016pruning}.
Using the process of minimizing the difference $d_{f_{k}^{i}, f_{k}^{j}}$ between the filters corresponding to a pair (which belongs to $Q_i$), we can increase the similarity between the filters that belong to the same filter pair. Thereby, one filter is pruned from each pair without affecting the model's performance much. The pruned model contains the reduced number of trainable parameters $W\textprime$,    
\begin{equation}
    W\textprime = W\backslash \{p_1, p_2, ...,p_k\},
    \label{eq:reducedmodel}
\end{equation}
where $p_1$, $p_2$, .. $p_k$ are the filters that are selected for pruning after optimization. Further, we re-train the network w.r.t. the reduced parameters $W\textprime$ to regain the classification performance. As we prune the redundant filters from the network, the information loss is minimum. Therefore, re-training (fine-tuning) makes the network to recover the loss incurred due to filter pruning.

\section{Experiments and Results}
\label{sec:experimental_results}
To demonstrate the significance of the proposed history based filter pruning method, we utilize four popular deep learning models, VGG-16 \cite{simonyan2014very}, ResNet-50, ResNet-56, and ResNet-110 \cite{he2016deep}. All the experiments are conducted on NVIDIA GTX 1080 Titan Xp GPU.  Through our experimental results, we observe that our method obtains state-of-the-art model compression results for all the above mentioned CNNs. Similar to \cite{singh2020leveraging}, the regularizer term $\lambda$ is set to $1$ for LeNet-5 and ResNet-50/56/110. However, we empirically observe that the value of $\lambda$ as $0.8$ gives better results for VGG-16. We prune $M \%$ of the filters from each convolutional layer simultaneously. The $M \%$ is considered as $10 \%$ for LeNet-5, VGG-16. Whereas, from ResNet-56, ResNet-110, and ResNet-50, we prune $2$, $4$, $8$ filters from each convolutional layer correspond to three blocks. We repeat this pruning process until there is a performance drop below a certain threshold, which is also a hyper-parameter. In our experiments, we set the threshold value $1\%, 2\%$, $2 \%$, and $3 \%$ for LeNet-5 \cite{lecun1998gradient}, VGG-16 \cite{simonyan2014very}, ResNet-56/110 \cite{he2016deep}, ResNet-50 \cite{he2016deep} models, respectively. Next, we discuss the datasets utilized for conducting the experiments and then we present a comprehensive results analysis and discussion.

\subsection{Datasets}
In this work, we use three popular and benchmark image classification datasets, namely MNIST, CIFAR-10, and ImageNet to conduct the experiments. 
\subsubsection{MNIST}
The MNIST dataset \cite{lecun2010mnist} consists the images of hand-written digits ranging from $0$ to $9$. This dataset has $60,000$ training images with $6,000$ training images per class and $10,000$ test images with $1,000$ test images per class. The dimension of the image is $28\times28\times1$. The LeNet-5 \cite{singh2020leveraging} is trained from scratch on MNIST. 

% Pruning results of LeNet-5 on MNIST is presented in the supplementary material.

\begin{table}[!t]
\centering
\caption{The pruning results of VGG-16 on CIFAR-10 dataset.
% We conduct experiments for three times and report the mean $\pm$ standard deviation.
The reported results are taken from the respective research article, except * which is reproduced. }
\resizebox{\columnwidth}{!}{%
\begin{tabular}{|l|l|l|l|}
\hline
Model                  & Top-1$\%$ & \begin{tabular}[c]{@{}l@{}}Remaining FLOPs \\ {(Drop \%)} \end{tabular}                                      &  \begin{tabular}[c]{@{}l@{}}Remaining \\ Parameters \\ {(Drop \%)} \end{tabular}   \\ \hline
VGG-16 \cite{simonyan2014very}  & 93.96 & 313.73M (0.0\%) & 14.98M  (0.0\%) \\  
$\ell_1$-norm \cite{li2016pruning}          & 93.40          & 206.00M (34.30\%)           & 5.40M (64.00\%)          \\ 
GM \cite{he2019filter}                              & 93.58           &   201.10M (35.90\%)          & -          \\ 
% PP \cite{}
VFP \cite{zhao2019variational}                          & 93.18          & 190.00M (39.10\%)            & 3.92M (73.30\%)          \\ 

Ayinde \etal \cite{ayinde2019redundant}                          & 93.67          & 186.67M (40.50\%)            & 3.28M (78.10\%)          \\

SSS \cite{huang2018data}                                & 93.02          & 183.13M (41.60\%)            & 3.93M  (73.80\%)          \\ 
% $\ell_1$-norm \cite{li2016pruning}                         & $2.06$          & 6.60            & 34.20          \\ \hline

FPEI \cite{wang2021filter} & 92.49
& 177.27M (43.45\%) & 3.30M (77.60\%) \\

GAL \cite{lin2019towards} & 90.73
& 171.89M (45.20\%) & 3.67M (82.20\%) \\

% Hinge \cite{li2020group} & 93.59 & 122.5M (60.93\%) & 11.99M (19.95\%) \\

% PFF \cite{meng2020pruning} & 92.85 &           90.47M (71.16\%) & 1.09M (92.66\%) \\

Chen \etal \cite{chen2020dynamical} & 92.90 & 118.70M (51.01\%) & 5.50M (63.28\%) \\
\textbf{HBFP-I (Ours)}               & 93.04  &  90.23M (71.21\%)  & 4.20M (71.80\%) \\ 

\textbf{HBFP-II (Ours)}                 & 92.54  & 75.05M (76.05\%)   & 3.50M (76.56\%) \\ 

HRank \cite{lin2020hrank} & 91.23 & 73.70M (76.50\%) & 1.78M (92.00\%) \\

% FO \cite{qin2018functionality}          & $1.85$          & 6.70            & 44.10          \\ \hline

\textbf{HBFP-III (Ours)}             & 92.30  & 62.30M (80.09\%)   &  2.90M (80.47\%) \\

CFP \cite{singh2020leveraging}$^*$          & 91.83         &     59.15M (81.14 \%)       &   2.80M (81.10\%)       \\ 
CFP \cite{singh2020leveraging}          & 92.98          &      56.70M (81.93\%)       &     -      \\

% CFP-II \cite{singh2020leveraging}$^*$          & 91.24         &     49.12M (84.34 \%)       &   2.3M (84.37\%)       \\ 

% Auto-balanced FP \cite{ding2018auto}          & $0.58$          & 7.13            & 81.39          \\ \hline

\textbf{HBFP-IV (Ours)}                 & 91.99  & 51.90M (83.42\%)   & 2.40M (83.77\%) \\ \hline

\end{tabular}
}
\label{tab:vgg_cifar10}
\end{table}

\subsubsection{CIFAR-10}
CIFAR-10 \cite{krizhevsky2009learning} is the most widely used tiny-scale image dataset which has images belonging to $10$ object categories. The dimension of the image is $32\times32\times3$. This dataset contains a total of $60,000$ images with $6,000$ images per class, out of which $50,000$ images (i.e., $5,000$ images per class) are used for both training and fine-tuning the network and the remaining $10,000$ images (i.e., $1,000$ images per class) are used for validating the network performance.

\subsubsection{ImageNet}
ImageNet \cite{deng2009imagenet} is a large-scale visual recognition image dataset consists of 1.2 Million training images, 50,000 validation images belong to 1,000 object categories. We have downsampled the image dimension from $256\times256\times3$ to $224\times224\times3$ to conduct the experiments.

\begin{figure*}[htp]
    \centering
    \includegraphics[width=0.9\textwidth]{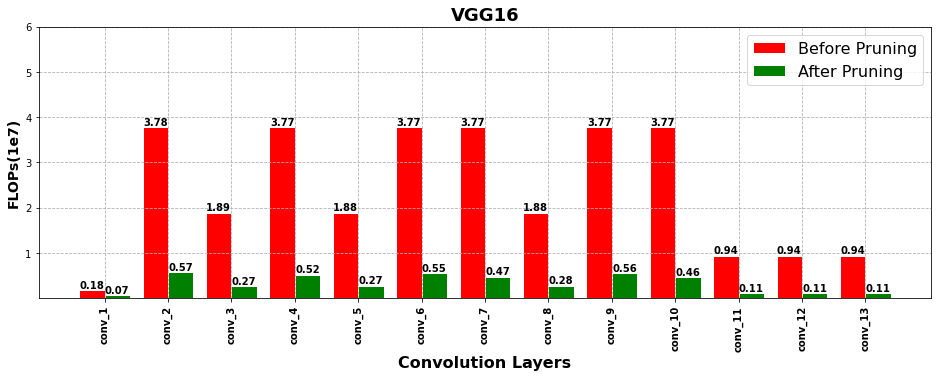}
    \caption{The total number of FLOPS involved in each convolutional layer before and after pruning for VGG-16 trained on CIFAR-10 using the proposed history based filter pruning approach. 
    % We can observe top-2 compression (FLOPs reduction) rates in convolutional layers $conv_1$ and $conv_8$. 
    }
    \label{fig:vgg_flops}
\end{figure*}

\begin{table}[!t]
\centering
\caption{The pruning results of ResNet-56/110 on CIFAR-10 dataset. }
\resizebox{\columnwidth}{!}{%
\begin{tabular}{|l|l|l|l|}
\hline
Model                  & Top-1$\%$ & Pruned FLOPs                                      &  \begin{tabular}[c]{@{}l@{}}Remaining \\ Parameters \\ {(Drop \%)} \end{tabular}   \\ \hline
ResNet-56 \cite{he2016deep}  & 93.26 & 125.49M (0.0\%) & 0.85M  (0.0\%) \\  
VFP \cite{zhao2019variational}                          & 92.26          & 96.60M (20.30\%)            &  0.67M (20.49\%)          \\

$\ell_1$-norm \cite{li2016pruning}          & 93.06         & 90.90M (27.60\%)           & 0.73M (14.10\%)          \\ 

Ayinde \etal \cite{ayinde2019redundant}          & 93.12         & 90.70M (27.90\%)           & 0.65M (23.70\%)          \\ 

NISP \cite{yu2018nisp}                              & 93.01           &   81.00M (35.50\%)          & 0.49M (42.40\%)          \\

\textbf{HBFP-I (Ours)}               & 92.42  &  70.81M (43.68\%)  & 0.48M (46.38\%) \\

AMC \cite{he2018amc} & 91.90 & 62.74M (50.00\%) & - \\

CP \cite{he2017channel} & 91.80 & 62.74M (50.00\%) & - \\

Chen \etal \cite{chen2020dynamical} & 93.10 & 62.70M (50.03\%) & 0.43M (49.04 \%) \\

He \etal \cite{he2019asymptotic} & 93.12 & 59.40M (50.03\%) & - \\

RUFP \cite{zhang2022rufp} & 93.17
& 53.70M (57.70\%) & - \\

GAL \cite{lin2019towards} & 90.36
& 49.99M (60.20\%) & 0.29M (65.90\%) \\

\textbf{HBFP-II (Ours)}                 & 92.25  & 49.22M (60.85\%)   & 0.33M (60.85\%) \\

% PFF \cite{meng2020pruning} & 

HRank \cite{lin2020hrank} & 90.72 & 32.52M (74.10\%) & 0.27M (68.10\%) \\

\textbf{HBFP-III (Ours)}             & 91.79  & 31.54M (74.91\%)   &  0.21M (74.90\%) \\ 

CFP \cite{singh2020leveraging}$^*$          & 91.37          &      31.54M (74.91\%)       &    0.21M (74.90\%)      \\ 

 CFP \cite{singh2020leveraging}          & 92.63         &      29.50M (76.59 \%)    &    3.40M (77.14\%) \\
 
\textbf{HBFP-IV (Ours)}                 & 91.42  & 27.10M (78.43\%)   & 0.19M (76.97\%) \\ \hline

ResNet-110 \cite{he2016deep}
& 93.50 & 252.89M (0.0\%) & 1.72M (0.0\%) \\

VFP \cite{zhao2019variational}                          & 92.96          & 160.70M (36.44\%)            &  1.01M (41.27\%)          \\

$\ell_1$-norm \cite{li2016pruning} & 93.30 & 155.0M (38.70\%) & 1.16M (32.60\%) \\
GAL \cite{lin2019towards} & 92.55 & 130.20M (48.50\%) & 0.95M (44.80\%)
\\

Ayinde \etal \cite{ayinde2019redundant} & 93.27 & 154.00M (39.10\%) & 1.13M (34.20\%) \\
GAL \cite{lin2019towards} & 92.55 & 130.20M (48.50\%) & 0.95M (44.80\%)
\\

Chen \etal \cite{chen2020dynamical} & 93.86 & 126.4M (50.00\%) & 0.88M (48.91\%)
\\
He \etal \cite{he2019asymptotic} & 93.1 & 121.1M (52.30\%) & -
\\

\textbf{HBFP-I (Ours)}             & 93.11  & 119.70M (52.69\%)   &  0.81M (52.66\%) \\

\textbf{HBFP-II (Ours)}                 & 92.91  & 98.90M (60.89\%)   & 0.67M (41.27\%) \\

\textbf{HBFP-III (Ours)}             & 92.83  & 80.20M (68.31\%)   &  0.54M (68.28\%) \\

HRank \cite{lin2020hrank} & 92.65 & 79.30M (68.60\%) & 0.53M (68.70\%) \\

\textbf{HBFP-IV (Ours)}                 & 91.96  & 63.30M (74.95\%)   & 0.43M (74.92\%) \\ \hline

\end{tabular}
}
\label{tab:resnet_results}
\end{table}

\subsection{LeNet-5 on MNIST}
We utilize LeNet-5 architecture which has two convolutional layers $Conv1$ and $Conv2$ with  20 and 50 filters, respectively, of spatial dimension $5\times5$. The first convolutional layer $Conv1$ is followed by a max-pooling layer $Max\_pool1$ with $2\times2$ filter which results in a feature map of dimension $12\times12\times20$. Similarly, the second convolutional layer $Conv2$ is followed by another max-pooling layer $Max\_pool2$ with $2\times2$ dimensional filter which results in a $4\times4\times50$ dimensional feature map. The feature map resulted by $Max\_pool2$ layer is flattened into a $800\times1$ dimensional feature vector which is given as input to the first Fully Connected (FC) layer $FC1$. The $FC1$ and $FC2$ layers have $500$ and $10$ neurons, respectively. The LeNet architecture corresponds to $4,31,080$ trainable parameters and 4.4M FLOPs.

We conduct the pruning experiment on LeNet-5 over the MNIST dataset using the proposed HBFP method. Training the network from scratch results in a $0.83\%$ base error. The comparison among the benchmark pruning methods for LeNet-5 is shown in Table \ref{tab:lenet_results}. Compared to the previous pruning methods, the proposed HBFP method achieves a higher reduction in the FLOPs, i.e., $97.98 \%$, however, still results in a less error rate, i.e., $1.4 \%$. Structured Sparsity Learning (SSL) method pruned $93.42\%$ FLOPs with $1\%$ error rate. From Table \ref{tab:lenet_results} (the rows corresponding to the proposed HBFP method), we can examine that the proposed method achieves better classification performance with a high percent of pruning compared to other methods. The previous work on Correlation Filter Pruning (CFP) by Singh \etal \cite{singh2020leveraging} has reported a similar FLOPs reduction, i.e., $97.98 \%$, however, their error rate is $1.77 \%$ which is quite high compared to our method. Singh \etal \cite{singh2020leveraging} reported that employing a regularizer (in the form of an optimization) reduces the information loss that occurs due to filter pruning. Motivated by this work, we also optimize the network to increase the similarity between the filters belonging to a redundant filter pair. In this process, we reproduce the results of CFP \cite{singh2020leveraging} for which we obtain the top-1 error rate as $2.61 \%$ with the same percent of reduction in the FLOPs \mbox{(row 10 of Table \ref{tab:lenet_results})}.

\subsection{VGG-16 on CIFAR-10}
In 2014, Simonyan \etal \cite{simonyan2014very} proposed a VGG-16 CNN model that received much attention due to improved performance over ImageNet Large Scale Visual Recognition Challenge (ILSVRC). The same architecture and settings are used in \cite{simonyan2014very} with few modifications such as batch normalization layer \cite{ioffe2015batch} is added after every convolutional layer. The VGG-16 consists of $1,49,82,474$ trainable parameters and 313.73M FLOPs. 

Training the VGG-16 network from scratch enables the model to achieve $93.96 \%$ top-1 accuracy on the CIFAR-10 object recognition dataset. The comparison among the state-of-the-art filter pruning methods for VGG-16 on CIFAR-10 available in the literature is performed in Table \ref{tab:vgg_cifar10}. The proposed method prunes $83.42 \%$ of FLOPs from VGG-16 which results in $91.99 \%$ of top-1 accuracy. The Geometric Median method proposed in \cite{he2019filter} reported $93.58 \%$ top-1 accuracy with a $35.9 \%$ reduction in the FLOPs. The recent works, such as HRank \cite{lin2020hrank} and Correlation Filter Pruning (CFP) \cite{singh2020leveraging} are able to prune $76.5 \%$ and $81.93 \%$ FLOPs with the error rate of $8.77\%$ and $7.02 \%$, respectively, while the proposed HBFP method is able to prune $83.42 \%$ FLOPs with an error rate of $8.01 \%$. The detailed comparison of pruning results for VGG-16 on the CIFAR-10 dataset is demonstrated in Table \ref{tab:vgg_cifar10}. The FLOPs in each convolutional layer before and after employing the proposed pruning method on VGG-16 over CIFAR-10 is illustrated in Fig. \ref{fig:vgg_flops}. 

\subsection{ResNet-56/110 on CIFAR-10}
We also use the deeper CNN models such as ResNet-56 and ResNet-110 \cite{he2016deep} to conduct the pruning experiments over the CIFAR-10 dataset using the proposed HBFP method. ResNet-56, ResNet-110 have three blocks of convolutional layers with $16$, $32$, and $64$ filters. Training these residual models (i.e., ResNet-56 and ResNet-110) with the same parameters as in \cite{he2016deep} produce $93.26 \%$ and $93.5 \%$ top-1 accuracies, respectively. 
The HBFP method prunes $2$ filters from the first block which has $16$ filters, $4$ filters from the second block which has $32$ filters, and $8$ filters from the third block which has $64$ filters in each iteration of the proposed method. From Table \ref{tab:resnet_results}, it is evident that the proposed pruning method produces state-of-the-art compression results for ResNet-56 and ResNet-110 on CIFAR-10 dataset. 

\textbf{ResNet-56:} As depicted in Table \ref{tab:resnet_results}, both AMC \cite{he2018amc} and CP \cite{he2017channel} methods have reduced $50.00 \%$ FLOPs while resulting in $8.1 \%$ and $8.2 \%$ error, respectively. The HRank \cite{lin2020hrank} prunes $74.1 \%$ FLOPs with $9.28 \%$ error. From Table \ref{tab:resnet_results}, it is clear that the proposed HBFP obtains top-1 accuracy $91.42 \%$ with high reduction of FLOPs ($78.43 \%$) compared to HRank \cite{lin2020hrank}. However, the proposed method obtains the high FLOPs reduction with comparable performance $91.42 \%$ as compared to CFP \cite{singh2020leveraging}. Moreover, we achieve a better performance using the proposed HBFP as compared to the reproduced results using CFP \cite{singh2020leveraging}.

\textbf{ResNet-110:} As per the results summarized in the lower part of Table \ref{tab:resnet_results}, the filter's $\ell_1$-norm based filter pruning method obtained $93.3\%$ top-1 accuracy by pruning $38.7\%$ of the FLOPs. The recent HRank \cite{lin2020hrank} method achieved $92.65\%$ top-1 performance with $68.6\%$ FLOPs reduction. From Table \ref{tab:resnet_results}, we can note that our method achieves a $74.95\%$ FLOPs by removing $74.92 \%$ of the trainable parameters, with a minimum loss of $1.54\%$ as compared to the baseline using ResNet-110 on CIFAR-10. Moreover, our HBFP-III performs better than HRank \cite{lin2020hrank} in terms of the accuracy with comparable FLOPs reduction.

\begin{table}[!t]
\centering
\caption{The pruning results of ResNet-50 on ImageNet. The results are arranged in the decreasing order of pruned FLOPs.}
% We conduct experiments for three times and report the mean $\pm$ standard deviation.
% The reported results are taken from the respective research article, except * which is reproduced. }
\resizebox{\columnwidth}{!}{%
\begin{tabular}{|l|l|l|l|}
\hline
Model                  & Top-1$\%$ &  \begin{tabular}[c]{@{}l@{}}Remaining FLOPs \\ {(Drop \%)} \end{tabular}                                      &  \begin{tabular}[c]{@{}l@{}}Remaining \\ Parameters \\ {(Drop \%)} \end{tabular}  \\ \hline
Baseline   & 74.86 & 3.83B (0.0\%) & 25.53M  (0.0\%) \\  
SSS-32 \cite{huang2018data}      & 74.18          & 2.82B (26.30\%)           & 18.60M (27.10\%)          \\ 
He \etal \cite{he2017channel}                              & 72.30           &   2.73B (28.70\%)          & -          \\ 
% PP \cite{}
GAL-0.5 \cite{lin2019towards}                          & 71.95          & 2.33B (39.10\%)            & 21.20M (16.90\%)          \\ 

SSS-26 \cite{huang2018data}                                & 71.82          & 2.33B  (39.10\%)            & 15.60M  (38.80\%)          \\ 

% HRank-I \cite{lin2020hrank} & 74.98 & 2.30B (39.9\%) & 16.15M (36.7\%) \\

GDP-0.6 \cite{lin2018accelerating} & 71.19 & 1.88B (50.90\%) & - \\

GAL-1 \cite{lin2019towards}                          & 69.88          & 1.58B (58.70\%)            & 14.60M (42.80\%) \\

GDP-0.5 \cite{lin2018accelerating} & 69.58 & 1.57B (59.00\%) & - \\

% HRank-II \cite{lin2020hrank} & 71.98 & 1.55B (59.5\%) & 13.77M (46.06\%) \\

\textbf{HBFP-I (Ours)}               & 69.89  &  1.13B (70.49\%)  & 9.06M (64.50\%) \\ 

ThinNet-50 \cite{luo2017thinet} & 68.42 & 1.10B (71.27\%) & 8.66M (66.07\%) \\

\textbf{HBFP-II (Ours)}                 & 69.46  & 1.06B (72.32\%)   & 8.68M (66.00\%) \\ 

\textbf{HBFP-III (Ours)}             & 69.28  & 0.99B (74.15\%)   &  8.38M (67.17\%) \\ 

HRank \cite{lin2020hrank} & 69.10 & 0.98B (74.40\%) & 8.27M (67.60\%) \\

\textbf{HBFP-IV (Ours)}                 & 69.17  & 0.94B (75.45\%)   & 8.09M (68.30\%) \\ 

\hline

\end{tabular}
}
\label{tab:resnet50_imagenet}
\end{table}

\subsection{Ablation Study}
In the below section, we provide the ablation study on the effect of the proposed regularizer.
\subsection{Effect of Regularizer}
\begin{figure}[!t]
    \centering
    \includegraphics[width=0.45\textwidth]{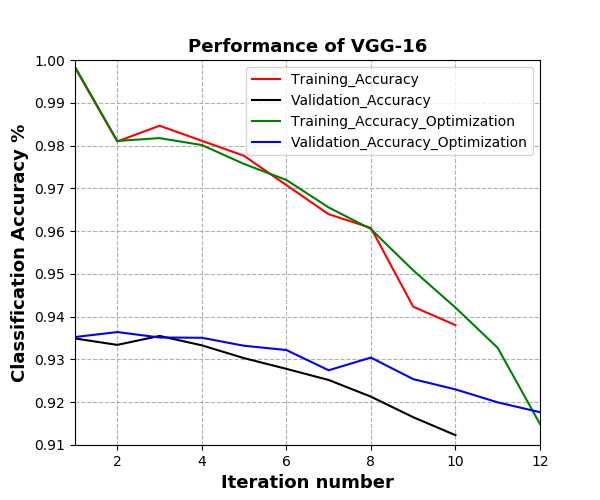}
    \caption{Illustrating the effect of pruning results obtained using the proposed pruning method with and without employing optimization. We can observe, introducing a custom regularizer to the objective increases the model's performance.}
    \label{fig:opt_woopt_error}
\end{figure}

To investigate the effect of the optimization step, we also conduct the experiments without employing the custom regularizer. We show the effect of the regularizer (optimization step) by comparing the classification results obtained for VGG-16 on CIFAR-10 using the HBFP method with and without employing the regularizer. From Fig. \ref{fig:opt_woopt_error}, it can be observed that increasing the similarity between the filters belong to a redundant filter pair using a regularizer and thereby training the network decreases the information loss that occurs due to filter pruning. The reason for minimum information loss is because the optimization step increases the redundancy level among the filters of a pair such that removal of one filter does not affect the performance. 

% We also perform similar experiments using other CNNs, the corresponding results are reported in the Supplementary material.

\subsection{ResNet-50 on ImageNet}
The MNIST and CIFAR-10 are medium-scale datasets consisting of low-resolution images. To demonstrate the efficacy of the proposed method over a large-scale dataset having high-resolution images, we also experiment with ImageNet dataset. The pruning results of ResNet-50 on ImageNet is shown in Table \ref{tab:resnet50_imagenet}. Training ResNet-50 with the same parameters as in \cite{he2016deep} except batch size produces the top-1 accuracy 74.86\%. We consider batch size as 32 due to computational resource limit (originally the ResNet-50 is trained with batch size 128). From Table \ref{tab:resnet50_imagenet}, we can observe that the proposed HBFP method results in 69.17 top-1 accuracy with 75.45\% and 68.3\% reduction in FLOPs and parameters, respectively. We can also observe that our method outperforms HRank \cite{lin2020hrank} (second last row) with increased percentage reduction in FLOPs and Parameters. Thus, it is observed that the HBFP model is also suitable for the large-scale datasets having high-resolution images.

% \subsection{Effect of Regularizer}
% To investigate the effect of the optimization step, we also conduct the experiments without employing the custom regularizer. We show the effect of the regularizer (optimization step) by comparing the classification results obtained for VGG-16 on CIFAR-10 using the HBFP method with and without employing the regularizer. From Fig. \ref{fig:opt_woopt_error}, it can be observed that increasing the similarity between the filters belong to a redundant filter pair using a regularizer and thereby training the network decreases the information loss that occurs due to filter pruning. The reason for minimum information loss is due to the fact that the optimization step increases the redundancy level among the filters of a pair such that removal of one filter does not affect the performance. We also perform similar experiments using other CNNs, the corresponding results are reported in the Supplementary material.

% \begin{figure}[htp]
%     \includegraphics[width=8cm]{graph2.png}
%     \caption{Bar plot for before and after flops of Resnet56)
%     }
%     \label{fig:proposed_method}
% \end{figure}

\section{Conclusion}
\label{sec:conclusion}
We propose a new filter pruning technique which uses the filters' information at every epoch during network training. The proposed History Based Filter Pruning (HBFP) method is able to prune a higher percent of convolution filters compared with state-of-the-art pruning methods. At the same time, the HBFP pruning can produce a less error rate. Eventually, it reduces the FLOPs available in LeNet-5 ($97.98 \%$), VGG-16 ($83.42 \%$), ResNet-56 ($78.43 \%$), ResNet-110 ($74.95\%$), ResNet-50 ($75.45 \%$) models. 
The main finding of this paper is to prune the filters that exhibit similar behavior throughout the network training as the removal of one such filter from a filter pair does not affect the model's performance greatly.
According to our study, employing the custom regularizer to the objective function also improves the classification results. We show the importance of the proposed pruning strategy through experiments, including LeNet-5 on MNIST, VGG-16/ResNet-56/ResNet-110 on CIFAR-10, and ResNet-50 on ImageNet. It is also observed that the proposed method is robust to low and high resolution images. One possible direction of future research is pruning the filters further by considering the similarity among the filters from different layers. 
%%%%%%%%% REFERENCES

\section{Acknowledgments}

We acknowledge the NVIDIA Corporation's support with the donation of GeForce Titan XP GPU, which is helpful in conducting the experiments of this research.

\bibliographystyle{IEEEtran}
\bibliography{egbib}

\end{document}